\documentclass[11pt,a4paper]{article}
\pdfoutput=1
\usepackage{hyperref}
\usepackage[hyperref]{emnlp2018}
\usepackage{times}
\usepackage{amssymb}
\usepackage{times}
\usepackage{import}
\usepackage{graphicx}
\usepackage{amsmath}

\aclfinalcopy 

\title{Joint Image Captioning and Question Answering}

\author{Jialin Wu\thanks{$^*$Equal contribution} \and  Zeyuan Hu$^*$ \and Raymond J. Mooney \\
  Department of Computer Science \\
  The University of Texas at Austin\\
  {\tt \{jialinwu, zeyuanhu, mooney\}@cs.utexas.edu} }

\date{}

\begin{document}
\maketitle
\begin{abstract}
Answering visual questions need acquire daily common knowledge and model the semantic connection among different parts in images, which is too difficult for VQA systems to learn from images with the only supervision from answers. Meanwhile, image captioning systems with beam search strategy tend to generate similar captions and fail to diversely describe images. To address the aforementioned issues, we present a system to have these two tasks compensate with each other, which is capable of jointly producing image captions and answering visual questions. In particular, we utilize question and image features to generate question-related captions and use the generated captions as additional features to provide new knowledge to the VQA system. For image captioning, our system attains more informative results in term of the relative improvements on VQA tasks as well as competitive results using automated metrics. Applying our system to the VQA tasks, our results on VQA v2 dataset achieve 65.8\% using generated captions and 69.1\% using annotated captions in validation set and 68.4\% in the test-standard set. Further, an ensemble of 10 models results in 69.7\% in the test-standard split.
\end{abstract}

\section{Introduction}
In recent years, visual question answering (VQA) \cite{antol2015vqa} and image captioning task \cite{chen2015microsoft} have been widely and separately studied in both computer vision and NLP communities. Most of the recent works \cite{anderson2017bottom,singh2018attention,lu2017knowing,rennie2017self,pedersoli2017areas} concentrate on designing attention modules to better gather image features for both tasks. Those attention modules help the systems learn to focus on potential relevant semantic parts of the images and improve performance to some extent. 

However, the qualities of the attention are not guaranteed since there is no direct supervision on them. The boundaries of the attended regions are often vague in the top-down attention modules which fail to filter out noisy irrelevant parts of images. Even though the bottom-up attention mechanism \cite{anderson2017bottom} ensures clear object boundaries from object detection, it is still questionable that whether we can attend to the accord semantic parts from multiple detected regions given the insufficient amount of detected object categories and the lack of supervision on the semantic connections among those objects. Fully understanding the images need acquire daily
common  knowledge  and  model  the  semantic connection  among  different  parts  in  images, which goes beyond what the attention modules can learn from images only. Therefore additional image descriptions can be a helpful common knowledge supplement to the attention modules. In fact, we find that captions are very useful to the VQA tasks. In VQA v2 validation split, we answer questions using the ground truth captions \textbf{without} images and achieve 59.6\% accuracy, which has already outperformed a large number of VQA systems using image features only.

\begin{figure*}[!t]
\centering
\includegraphics[width=0.9\linewidth,trim={1.5cm 2.5cm 6.5cm 4cm},clip]{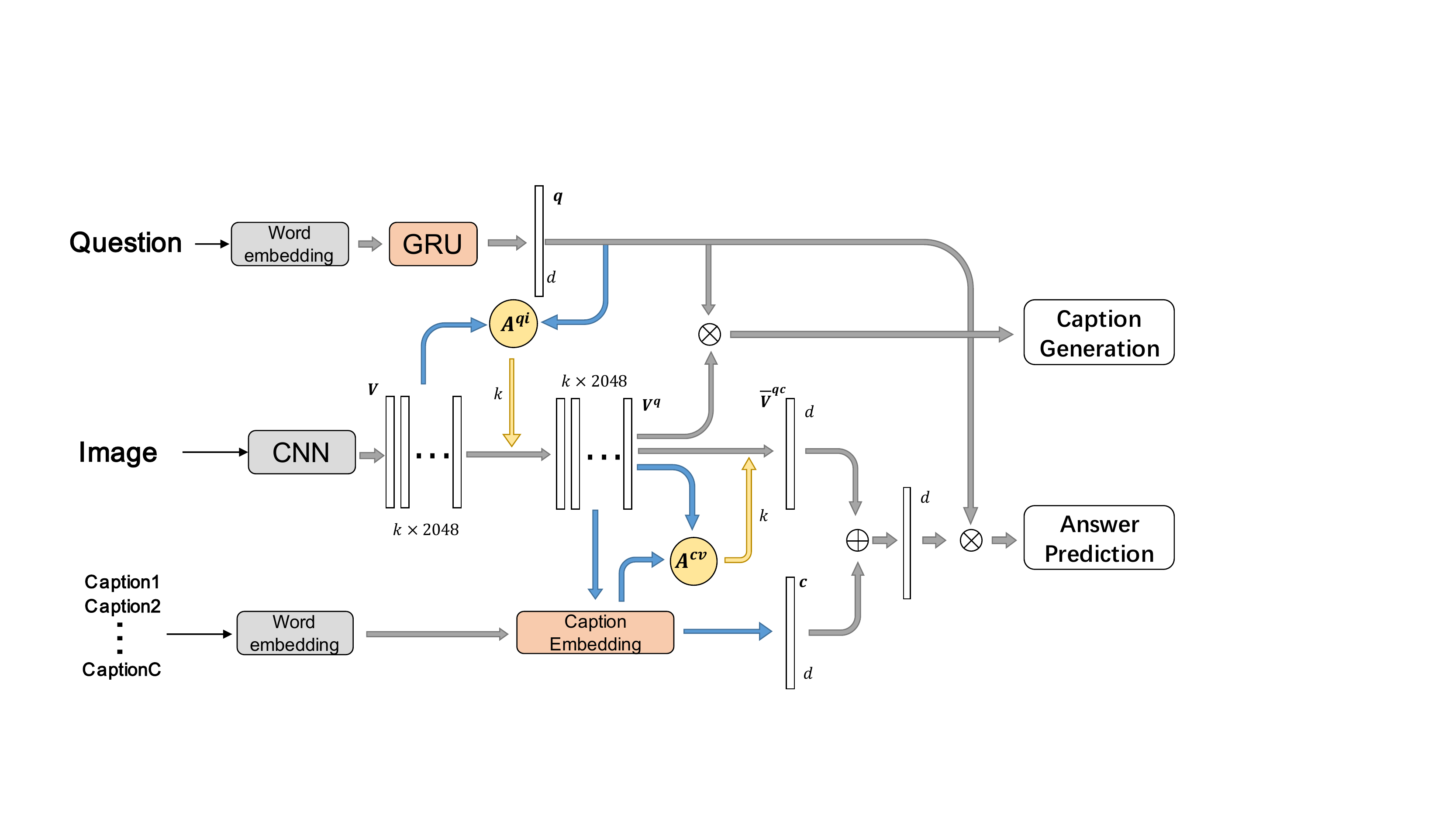}
\caption{Overall structure of our joint image captioning and VQA system. Our system takes questions, images, and captions as inputs and uses questions and images' joint representation to generate question related captions. Then we use the joint representation of the three inputs to predict answers. The numbers around the rectangle indicate dimensions, $\otimes$ denotes element-wise multiplication and $\oplus$ denotes element-wise addition. Blue arrows denote $fc$ with learnable parameters and yellow arrows denote attention embedding.}
\label{fig:overall_structure}
\end{figure*}

On the other hand, image captioning systems with beam search strategy tend to generate short and general captions based on the most significant objects or scenes. Therefore the captions are usually less informative in that they fail to diversely describe the images and build complex relationships among different parts of images. To obtain more specific and diverse captions, more heuristic could be helpful if provided. Since VQA process require to be aware of various aspects in the images, visual questions, as the additional heuristics, can potentially help the image captioner explore richer image content. To quantitatively measure informativeness, besides the automated evaluation metrics, we propose to use the relative improvements on the VQA accuracies when systems take the generated captions into consideration.

In this work, we propose to jointly generate question-related captions and answer visual questions. We demonstrate that these two tasks can be a good complement to each other. On the one hand VQA task provides the image captioners more heuristic during captioning. On the other hand, the captioning task feeds more common knowledge to the VQA system. Specifically, we utilize the joint representation of questions and images to generate question-related captions which serve as additional inputs to the VQA systems as hints. Furthermore, creating these captions reduces the risk of learning from questions bias \cite{li2018tell} and ignoring the image content when high accuracy sometimes can be already achieved from the questions solely. Meanwhile, questions, as novel heuristics, inspire the captioning systems to generate captions that are question-related and helpful to the VQA process. To automatically choose annotated question-related captions in training phase as supervision, we propose an online algorithm which selects the captions that maximize the inner-products between gradients from the captions and answers to the images and questions' joint representation. 

For the evaluation of the joint system, we first evaluate the benefits from additional caption inputs in the VQA process. Empirically, we observe a huge improvements on the answers accuracy over the BUTD\cite{anderson2017bottom} baseline model in the VQA v2 validation splits \cite{antol2015vqa}, from 63.2\% to 69.1\% with annotated captions from COCO dataset \cite{chen2015microsoft}  and 65.8\% with generated captions. And our single model achieves 68.4\% in the test-standard split. Furthermore, an ensemble of 10 models results in 69.7\% in the test-standard split. For the image captioning task, we show that our generated captions can not only achieve promising results under the standard caption evaluation metrics but also provide more informative descriptions in terms of relative improvement on the VQA task.

\section{Approach}
We first describe the overall structure of our joint system in Sec. \ref{sec:overview} and explain the feature representation as the foundations in Sec.\ref{sec:feat_repr}. Then, the VQA sub-system is introduced concretely in Sec. \ref{sec:vqa}, which takes advantage of image captions to better understand the images and thus achieve better results in VQA task. In sec. \ref{sec:ic}, we explain the image captioning sub-system which takes the question features as additional heuristics and generates more informative captions. Finally, the training details are provided in Sec. \ref{sec:training}.

\subsection{Overview}
\label{sec:overview}
We introduce the overall structure of our joint system. As illustrated as Fig. \ref{fig:overall_structure}, the system firstly produces images features $\textbf{V}=\{\textbf{v}_1, \textbf{v}_2, ..., \textbf{v}_K\}$ using bottom-up attention strategies, question features $\textbf{q}$ with standard GRU, caption features $\textbf{c}$ with our caption embedding module detailed in Sec. \ref{sec:feat_repr}. After that,  features $\textbf{q}$ and $\textbf{c}$ are used to generate the visual attention $\textbf{A}^{qv}$, $\textbf{A}^{cv}$ to weight the images' feature set $\textbf{V}$, producing question-caption-attended image features $\textbf{v}^{qc}$. We then sum up the $\textbf{v}^{qc}$ with caption features $\textbf{c}$ together and further perform element-wise multiplication with question features $\textbf{q}$. Meanwhile we use $\textbf{V}^q$ as input features to the image captioning sub system to generate question-related captions.

\subsection{Feature representation}
\label{sec:feat_repr}
We concretely introduce the representation of the images, questions and captions which serves as the foundation of the joint system. And we adopt $f(x)$ to denote a $fc$ layers $f(x) = LReLU(Wx + b)$ with input features $x$ and ignore the notation of weights and biases within the layers for simplicity, and those $fc$ layers do not share weights. The $LReLU$ denotes Leaky ReLU \cite{he2015delving}.\\

\noindent\textbf{Image and Question Embedding}\\
We adopt bottom-up attention \cite{anderson2017bottom} from object detection task to provide salient regions with clear boundaries in the images. In particular, we use Faster R-CNN head \cite{girshick2015fast} in conjunction with ResNet-101 base network \cite{he2016deep}. To generate an output set of image features $\textbf{V}$, we take the  final  outputs  of  the  model  and  perform  non-maximum suppression (NMS) for each object category using an IoU threshold $0.7$. And a fixed number of $36$ detected objects are extracted as the image features as suggested in \cite{teney2017tips}.

For the question embedding, we use a standard GRU with $1280$ hidden units inside and extract the output of hidden unit at final time step as the question features $\textbf{q}$.
And following \cite{anderson2017bottom}, the question features $\textbf{q}$ and image feature set $\textbf{V}$ are further embedded together to produce question-attended image feature set $\textbf{V}^q$ via the question visual attention $\textbf{A}^{qv}$ as illustrated in Fig. \ref{fig:overall_structure}.\\ 

\noindent\textbf{Caption embedding}\\
We present our novel caption embedding module in this section, which takes the question-attended image feature set $\textbf{V}^q$, question features $\textbf{q}$ and $C$ captions $\textbf{W}^c_i = \{w^c_{i, 1}, w^c_{i, 2}, ..., w^c_{i,T}\}$, where $T$ denotes the length of the captions and $i=1,...,C$ is the caption index, as inputs and generate the caption features $\textbf{c}$. 

\begin{figure}[h]
\centering
\includegraphics[width=\linewidth,trim={0cm 8.5cm 22.5cm 0cm},clip]{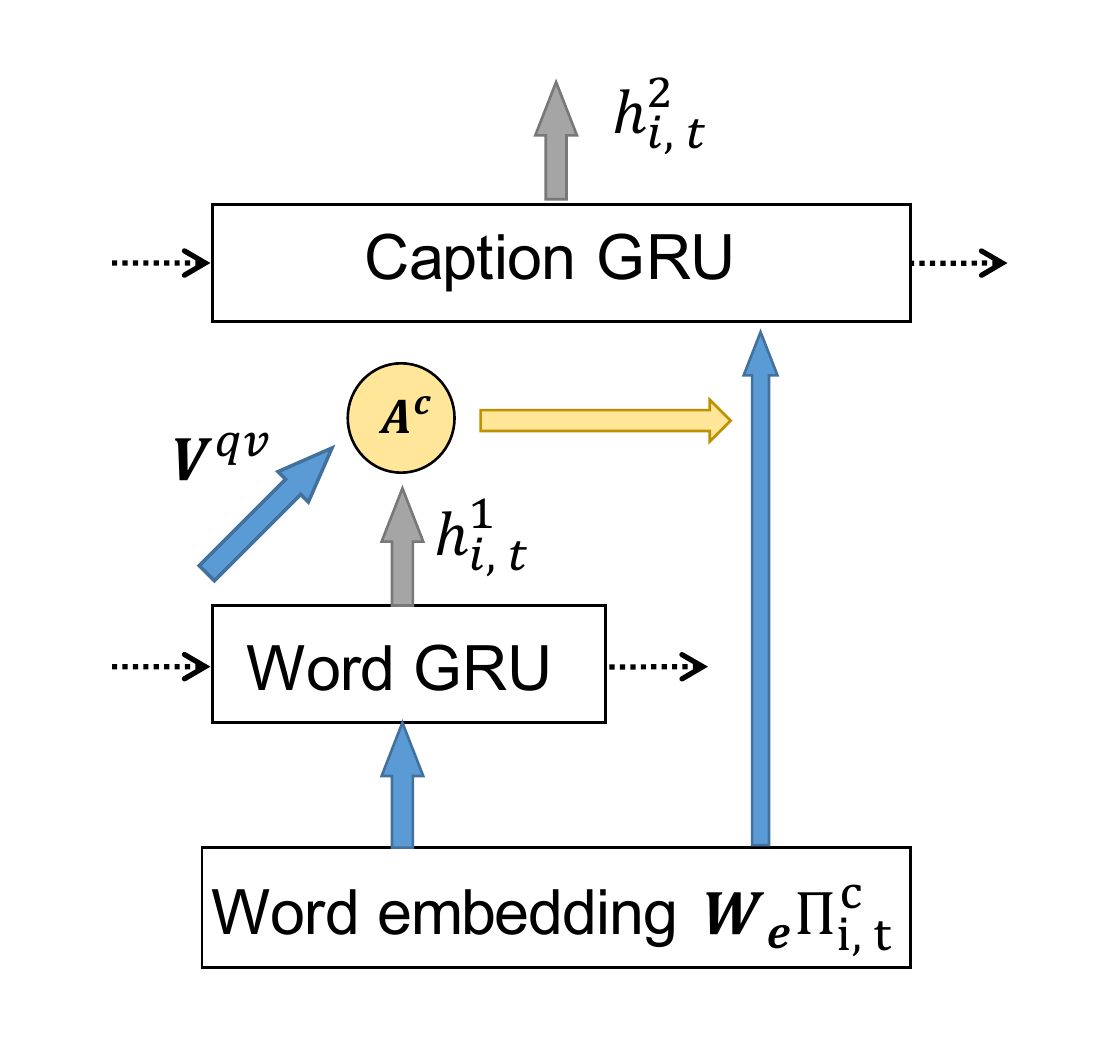}
\caption{Overview of the caption embedding module. The Word GRU is used to generate attention to identify the important and related words in each caption and Caption GRU outputs the caption embeddings. And we use question-attended image features $\textbf{V}^{qv}$ to model the attention. Blue arrows denote $fc$ with learnable parameters and yellow arrows denote attention embedding.}
\label{fig:caption_module}
\end{figure}

The goals of the caption module are to (1) provide additional descriptions to help the VQA system better understand the images by serving as a complement to the insufficiency of detected objects and the lack of common knowledge;(2) provide additional clues to models the semantic connection of the different parts in images from the bottom-up attention. 

To achieve that, as illustrated in Fig. \ref{fig:caption_module}, a Word GRU are first adopted to sequentially encode the words embedding in caption $\textbf{W}^c_i$ at each time step as $h^1_{i,t}$. Then, we design a caption attention module $\textbf{A}^c$ which utilize the question-attended image feature set $\textbf{V}^{q}$ and $h^1_{i,t}$ to generate new attentions on the current word to identify relevancy in an online fashion. Specifically, for each input word $\textbf{W}^c_i$, we use a standard GRU as the Word GRU to encode the words embedding $\Pi^c_{i,t}$ in Eq. \ref{eq:caption_word_attention}, and feed the outputs and $\textbf{V}^{q}$ to the attention module $\textbf{A}^c$ as shown in Eq. \ref{eq:caption_attention}. 
\begin{align}
    h^1_{i,t} &= GRU( \textbf{W}_e \Pi^c_{i,t},~ h^1_{i, t-1}) \\
    \label{eq:caption_word_attention}
    \overline{\textbf{v}}^{q} &= \sum_{k = 1}^{K} \textbf{v}_k^{q}\\
    a^{c}_{i,t} &= h^1_{i,t} \circ f(\overline{\textbf{v}}^{q}) \\
    \label{eq:caption_attention}
    \alpha^c_{i, t} &= \sigma(a^c_{i, t})
\end{align}
where $\sigma$ denotes sigmoid function, $K$ is the number of the bottom up attention, $\textbf{W}_e$ is the word embedding matrix and $\Pi^c_{i,t}$ is the one-hot embedding for $w^c_{i,t}$

After that, the attended words in captions are used to produce the final caption representation in Eq. \ref{eq:attended_caption_feed_foreward} via the Caption GRU. Since the goal is to gather more information, we perform element-wise max pooling within all input caption representation $\textbf{c}_i$ in Eq. \ref{eq:caption:maxp}.
\begin{align} 
\label{eq:attended_caption_feed_foreward}
    h^2_{i,t} &= GRU( \alpha^c_{i, t}\textbf{W}_e \Pi^c_{i,t}, ~ h^2_{i,t-1})\\
    \textbf{c}_i &= f(h^2_{i,T})\\
    \label{eq:caption:maxp}
    \textbf{c} &= max(\textbf{c}_i)
\end{align}
where $max$ denotes the element-wise max pooling over all captions $\textbf{c}$ of the images.
\subsection{VQA sub-system}
\label{sec:vqa}
We elaborate our VQA sub-system in details in this section. For the purpose of better mining semantic connection of different parts in images, our VQA system takes additional caption embeddings $\textbf{c}$ as inputs to generate caption-attend image attention $\alpha^{cv}$ in Eq.\ref{eq:image_caption_attention} and produce question-caption-attended image feature $\overline{\textbf{v}}^{qc}$ in Eq. \ref{eq:question-caption-attended}

\begin{align}
    a^{cv}_{i} &= f(f(\textbf{c})\circ f(\overline{v}^q_i))\\
    \label{eq:image_caption_attention}
    \alpha^{cv}_{i} &=  softmax(a^{cv}_{c,i})\\
    \label{eq:question-caption-attended}
    \overline{\textbf{v}}^{qc} &=  \sum_{i}  \textbf{v}^{q}_{i} \alpha^{cv}_{i}
\end{align}

To better incorporate knowledge from captions in the VQA's reasoning process, we also sums up the caption features $\textbf{c}$ with the joint attended image features $\overline{\textbf{v}}^{qc}$ and then element-wisely multiplies with the question features as shown in Eq. \ref{eq:vqa_repr}
\begin{align}
\label{eq:vqa_repr}
    \textbf{h} &= \textbf{q} \circ (f(\overline{\textbf{v}}^{qc}) + f(\textbf{c}))\\
    \hat{s} &= \sigma(f(\textbf{h}))
\end{align}

We frame the answer prediction task as a multi-label classification problem. In particular, we adopt the soft scores, which are in line with the evaluation metric, as labels to supervise the sigmoid-normalized predictions as shown in Eq. \ref{eq:vqa_loss}. In case of multiple feasible answers, the soft scores can capture the occasional uncertainty in ground truth annotations. As suggested in \cite{teney2017tips}, we collect the candidate answers that appears more than $8$ times in the training splits which results in $3129$ answer candidates. 
\begin{equation}
    \mathcal{L}^{vqa} = -\sum^{M}_{i=1}\sum^{N}_{j=1} s_{i,j}\log\hat{s}_{i,j} + (1-s_{i,j})\log(1-\hat{s}_{i,j})
    \label{eq:vqa_loss}
\end{equation}
where the indices $i$, $j$ run respectively over the $M$ training questions and
$N$ candidate answers and the $s$ are the aforementioned soft answer scores.

\subsection{Image Captioning sub-system}
\label{sec:ic}
We adopt the same image captioning module from \cite{anderson2017bottom} to jointly take advantage of the bottom-up and top-down attention. For the more detail module structure, refer to \cite{anderson2017bottom}. The key difference between our scheme and theirs lies on the input features and the caption supervision. Specifically, we feed the question-attended images features $\textbf{V}^q$ as inputs for the caption generation and only use the question-related annotated captions as supervision. During the training process, we compute the following cross entropy loss for each caption indexed by $j$ of the images in Eq. \ref{eq:caption_loss} and back propagate the gradients only from the most related caption detailed in next sub section.\\
\begin{equation}
    \mathcal{L}^c_{j} = -\sum^T_{t=1}\log(p(y_t|y_{t-1}))
    \label{eq:caption_loss}
\end{equation}
\\

\noindent\textbf{Selecting relevant captions for training}\\
Because our goal is to generate question-related captions, we need provide the image captioning system relevant captions as supervision in training phase. To achieve that, some word similarities based offline methods \cite{li2018vqa} have been proposed, however, those offline methods are not capable of being aware of the semantic connection among images parts and thus take more risks of incorrectly understanding the images. To address this issue, we propose an online relevant captions selection scheme which guarantees our system to update with a shared descent direction \cite{wu2018dynamic} in both the VQA parts and the image captioning parts, ensuring the consistency of image captioning module and the VQA module in the optimization process.

Specifically, we frame the caption selection problem as following, where the $j$ is the selected caption index. We require the inner product of the current gradients from the VQA and captioning loss to be greater than a constant $\xi$ and select a caption which maximizes that inner products. 
\begin{equation}
\begin{split}
    \underset{j}{\arg \max}& \sum_{i} \left( \frac{\partial \mathcal{L}^{vqa}}{\partial \textbf{v}^{q}_i}\right) ^T\frac{\partial \mathcal{L}^{c}_j}{\partial \textbf{v}^{q}_i} \\
    s.t.~~~~~&\sum_{i} \left( \frac{\partial \mathcal{L}^{vqa}}{\partial \textbf{v}^{q}_i}\right) ^T\frac{\partial \mathcal{L}^{c}_j}{\partial \textbf{v}^{q}_i} > \xi \\
    \label{pro:caption_selection}
\end{split}
\end{equation}

Therefore, given the solution of the problem \ref{pro:caption_selection} $j^{\star}$, the final loss of the multi-task learning is the sum of the VQA loss and the captioning loss of the selected captions as shown in Eq. \ref{eq:total_loss}. If the problem \ref{pro:caption_selection} has no feasible solution, we will ignore the caption loss. \\
\begin{equation}
    \mathcal{L} = \mathcal{L}^{vqa} + \mathcal{L}^c_{j^{\star}}
    \label{eq:total_loss}
\end{equation}

\noindent\textbf{Caption Evaluation on the informativeness}\\
Most of automated caption evaluation metrics are based on the statistical analysis on the word level language models ($e.g.$ BLEU, METEOR, $etc$) from machine translation task. However, different with MT, image captioning task aim at generating informative descriptions. Therefore, in our case, we propose a new metric that measure the informativeness according to the additional relative performance improvements on the VQA task, where much common knowledge is required.

Formally, we define the metric as 
\begin{equation}
    Info. = \frac{\hat{s} - \hat{s}_0}{\hat{s}_0}
    \label{eq:ic_metric}
\end{equation}
where $\hat{s}_0$ denotes the score of the VQA sub-system with input caption features $\textbf{c}$ manually set to zeros and $\hat{s}_0$ denotes the VQA score with captions as additional features.
\subsection{Training}
\label{sec:training}
We train our joint system using AdaMax optimizer with batch size $384$ as suggested in \cite{teney2017tips}. And we split out the official validation set of the VQA v2 dataset for monitoring to tune the initial learning rate and figure out which number of epochs yielding the highest overall VQA scores. We find that training the joint model 20 epochs will be sufficient and more training epoch may lead to overfitting, resulting in sightly drop of the performance. 

To simplify our training process, we firstly extract the bottom up detection attention as image features $\textbf{V}$. Unlike \cite{anderson2017bottom}, we don't require any other training data from other dataset. We initialize the training process with annotated captions from COCO dataset and pre-train our system for 20 epochs with the final loss in Eq. \ref{eq:total_loss}. After that, we generate the captions using our current system for all question image pairs in the COCO's train, validation and test sets. Finally, we finetune our system using the generated captions with 0.25$\times$ learning rate for 10 epoch.

\begin{table*}[!t]
\centering
\begin{tabular}{l|ccc|c}
\hline
                   & \multicolumn{4}{c}{Test-standard} \\\hline
                       & Yes/No  & Num   & Other  &All\\ \hline\hline
Prior \cite{goyal2017making} &61.20 & 0.36 &1.17 & 25.98 \\
Language-only \cite{goyal2017making} & 67.01 &31.55 & 27.37 & 44.26\\
MCB \cite{fukui2016multimodal} & 78.82 & 38.28 & 53.36 & 62.27 \\
BUTD \cite{anderson2017bottom}   & 82.20    & 43.90  & 56.26 & 65.32   \\
VQA-E \cite{anderson2017bottom}   & 83.22 & 43.58 & 56.79 & 66.31   \\
Beyond Bilinear \cite{yu2018beyond} & 84.50 &  45.39 & 59.01 & 68.09\\
ours-single & \textbf{84.69}   &  \textbf{46.75}  &    \textbf{59.30}     & \textbf{68.37}    \\
ours-ensemble-10  &\textbf{86.15}   &  \textbf{47.41} &  \textbf{60.41}   & \textbf{69.66}    \\\hline
\end{tabular}
\caption{Comparison of our results in VQA task with the state-of-the-art VQA methods on validation set and test-standard set. Accuracies in percentage (\%) are reported.}
\label{tab:vqa_compare}
\end{table*}
\section{Experiments}
We perform extensive experiments to evaluate our joint system in both VQA task and image captioning task. 
\subsection{Datasets}
\noindent\textbf{Image Captioning dataset}\\
We use the MSCOCO 2014 dataset for the image caption sub-system. In training stage, we use the dataset's official configuration but blind the Karpathy test split.

Following \cite{anderson2017bottom}, we perform only minimal text pre-processing, converting all sentences to lower case, tokenizing on white space, and filtering words that do not occur at least five times. For evaluation, we first use the standard automatic evaluation metrics, namely SPICE\cite{spice2016}, CIDEr \cite{vedantam2015cider}, METEOR \cite{banerjee2005meteor}, ROUGE-L\cite{lin2004rouge}, and BLEU\cite{Papineni:2002:BMA:1073083.1073135}. Further, we propose to measure the informativeness via the relative improvements on VQA task as shown in Eq. \ref{eq:ic_metric} as it requires much additional information.\\

\noindent\textbf{VQA dataset}\\
We use the VQA v2.0 dataset \cite{antol2015vqa} for the evaluation of our proposed joint system, where the answers are balanced in order to minimize the effectiveness of learning dataset priors. This dataset are used in VQA 2018 challenge and contains over 1.1M question from the images of MSCOCO 2015 dataset \cite{chen2015microsoft}.

Similar to the captions' data pre-processing, we also perform standard text preprocessing and tokenization. Following \cite{anderson2017bottom}, questions  are  trimmed to a maximum  of $14$ words. To evaluate answer qualities, we report accuracies using the official VQA metric, considering the occasional disagreement between annotators for the ground truth answers.

\subsection{Results on VQA task}
We report the experimental results of the VQA task and compare our results with the state-of-the-art methods. 
As demonstrated in Table. \ref{tab:vqa_compare}, our system outperforms the state-of-the-art systems which indicates the effectiveness of including caption features as additional inputs. In particular, we observe that our single model outperform other methods especially in 'Num' and 'Other' categories. The reasons are that the additional captions are capable of providing more numerical clues for infering 'Num' type questions and more common knowledge for answering the 'Other' type questions. Furthermore, an ensemble of 10 single models  results in 69.7\% in the test-standard split.\\

\noindent\textbf{Comparison between using the generated and annotated captions}\\
We then analyze the difference between generated and annotated captions. As demonstrated in Table. \ref{tab:gen_anno_compare}, our system gains about 6\% improvement from annotated captions and 2.5\% improvement from generated captions in the validation split. These evidence indicates the insufficiency of directly answering question from a limited number of detection and the necessity of incorporating additional knowledge from the images. And the generated captions are still not informative enough compared to the human annotated ones.\\
\begin{table}[h]
\centering
\begin{tabular}{l|c}
\hline
                  & Validation \\ \hline
BUTD \cite{anderson2017bottom} &  63.2\\
ours with BUTD captions &    64.6\\
ours with our generated captions &    65.8\\
ours with annotated captions &    \textbf{69.1} \\\hline
\end{tabular}
\caption{Comparison of the performance of using generated and annotated captions. Both of them provide large improvements to the baseline model. However, there are still a huge gap between generated captions and annotated captions.}
\label{tab:gen_anno_compare}
\end{table}

\begin{figure*}[!t]
\centering
\includegraphics[width=0.9\linewidth,trim={1.5cm 2cm 4cm 0cm},clip]{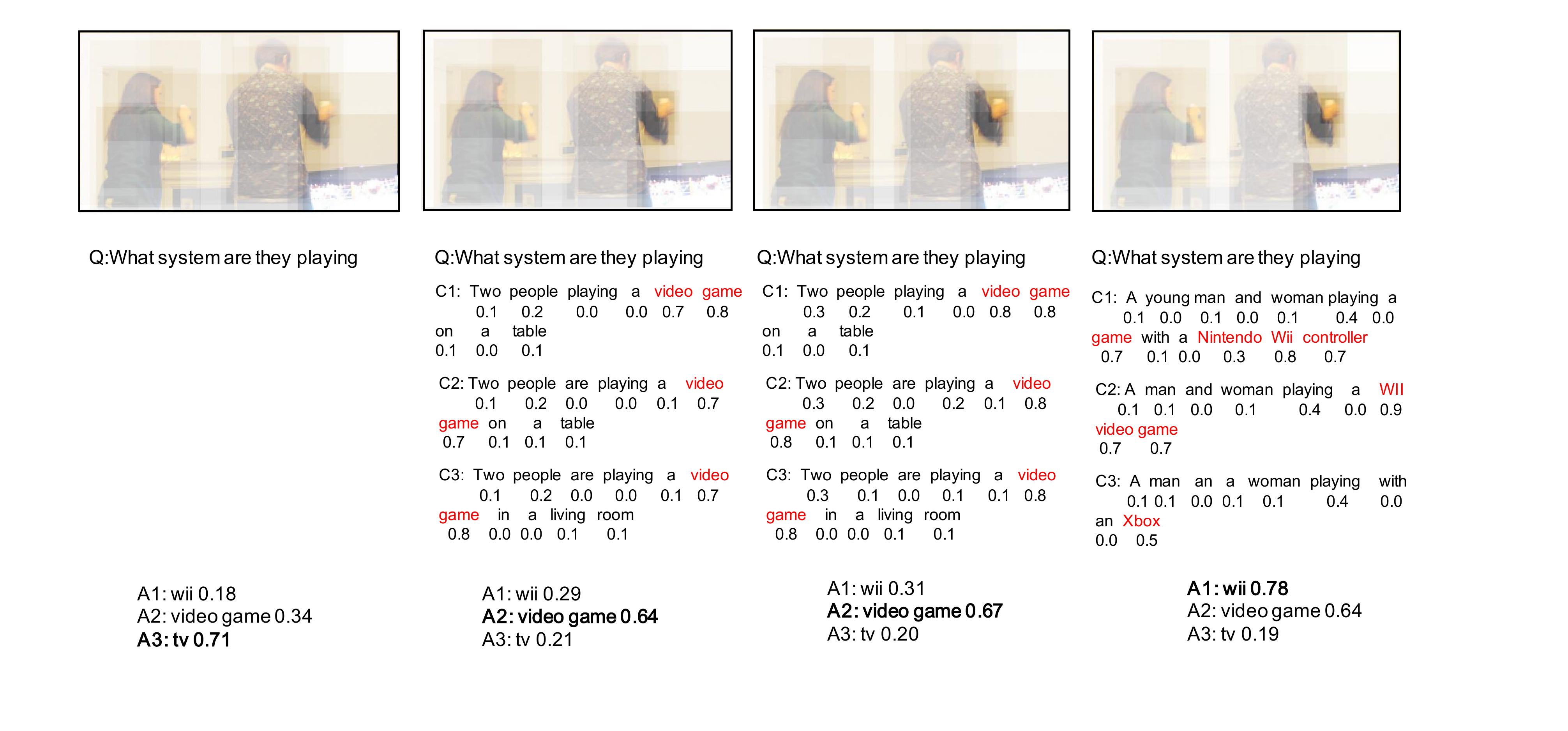}
\caption{Example of our joint system. The answers' scores in the questions are that $wii$ full score 1, video games score 0.3 and tv score 0.  Columns from left to right are BUTD, $w/o \& w \mathcal{SC}$ using generated captions, with annotated captions.}
\label{fig:attention_analysis}
\end{figure*}

\noindent\textbf{Ablation study on the semantic connection modeling}\\
In this section, we quantitatively analyze the effectiveness of incorporating captions to model the relationship besides the advantage of providing additional knowledge. The without relationship modeling methods is that we only use the caption features as inputs but don't involve them in the visual attention parts ($i.e.$ we don't compute $\textbf{A}^{cv}$). As demonstrated in Table. \ref{tab:relationship}, we observe above 0.5\% improvements are gained from adopting caption features to model the attention of images feature $\textbf{V}$ in both validation and test-standard splits. We use $w~\mathcal{SC}$ to indicate with semantic connection modeling modeling and $w/o~\mathcal{SC}$ to indicate without.\\ 
\begin{table}[h]
\centering
\begin{tabular}{l|cccc}
\hline
                  & \multicolumn{4}{c}{Test-standard} \\\hline
                  & All    & Yes/No  & Num   & Other  \\ \hline\hline
BUTD              & 65.3   & 82.2    & 43.9  & 56.3  \\
ours ($w/o~\mathcal{SC}$) & 67.4   & 84.0 & 44.5 & 57.9\\
ours ($w~\mathcal{SC}$) & 68.4  &  84.7 & 46.8 &  59.3     \\\hline
\end{tabular}
\caption{Evaluation of the effectiveness of semantic connection modeling in test-standard split. Accuracies in percentage (\%) are reported.}
\label{tab:relationship}
\end{table}

\begin{table}[h]
\centering
\begin{tabular}{l|cccc}
\hline
                  & \multicolumn{4}{c}{Validation} \\  \hline
                  & All   & Yes/No  & Num  & Other   \\\hline\hline
BUTD  & 63.2  & 80.3    & 42.8 & 55.8   \\
ours ($w/o~\mathcal{SC}$)  & 65.2  &  82.1 & 43.6 & 55.8\\
ours ($w~\mathcal{SC}$) &  65.8   &  82.6&  43.9& 56.4\\\hline
\end{tabular}
\caption{Evaluation of the effectiveness of semantic connection modeling in validation split. Accuracies in percentage (\%) are reported.}
\label{tab:relationship1}
\end{table}

\noindent\textbf{Qualitative Analysis}\\
We qualitatively analyze the attention inside the caption embedding module and the joint attention on the image features. As illustrated in Fig. \ref{fig:attention_analysis}, the attended image regions and attended caption words are visualized to confirm the correctness of the attended areas. Specifically, we observe that our system is capable of concentrating on the related objects ($i.e.$ the human hands and the TV monitor) and meaningful words in captions ($i.e.$ $wii$ in annotated captions and $video ~~game$ in generated captions) where BUTD baseline model \cite{anderson2017bottom} failed. And with semantic connection modeling, our system gains more confidence about its reasoning. As a results, the with ground truth captions, our system get full score by focusing on the $wii$ from the captions, and even with generated ones (with and without $\mathcal{SC}$), the system can get partial credits thanks to the $video ~~game$ in the captions. The reason is that captions provides additional common knowledge which is hard for the VQA system to directly learn from the images and those knowledge not only direct provide clues for the question answering process but also help the VQA system better understand the semantic connection among different images parts.

\begin{table*}[t]
\centering
\begin{tabular}{l|ccccccc}
\hline
     & \multicolumn{6}{c}{cross entropy loss}             \\\hline
     & BLEU-1 & BLEU-4 & METEOR & ROUGE-L & CIDEr & SPICE & Info.\\\hline\hline
BUTP & 77.2   & 36.2   & 27.0   & 56.4    & 113.5 & 20.3  & 2.22\%\\
ours & 77.4   & 36.7   & 25.8   & 56.4    & 110.2 & 20.8  & \textbf{4.11}\%\\\hline
\end{tabular}
\caption{Comparison of the image captioning task on the MSCOCO Karpathy test split. Our captioner obtains competitive results in automated evaluation metric and more informative compations in terms of relative improvements on VQA task.}
\label{tab:caption_comparison}
\end{table*}

\subsection{Results on Image captioning task}
In this section, we evaluate our joint system on image captioning task. In particular, we not only compare our system with the state-of-the-art systems via standard automated metrics, but also demonstrate that our generated question-related captions are more informative as they provide more benefits for the VQA task, using the measurements mentioned in Eq. \ref{eq:ic_metric}.

In Table. \ref{tab:caption_comparison}, we firstly report the standard automated caption metric score and compare them with the BUTD \cite{anderson2017bottom} model. When determine the captions for the images, we picked the most likely question-related captions based on the problem \ref{pro:caption_selection}. Our system, though only select a fraction of annotated caption as supervision, is capable of achieving similar standard scores. However, in term of the informativeness, we observe a large improvement on relative VQA accuracies from $2.22 \%$ to $4.11\%$. The improvements indicate that importance and necessity of adopting question features as inputs to the image captioning sub-system and more question-specific caption as supervision.

\section{Related Work}
\noindent\textbf{Visual question answering} \\
Recently, a large mount of attention-based deep learning methods have been proposed for VQA task including top-down attention methods \cite{gao2015you,Malinowski_2015_ICCV,ren2015exploring} and bottom up methods \cite{anderson2017bottom,li2018vqa} to incorporate different semantic parts in images. Specifically, the image features from pre-trained CNN, the question features from RNNs are combined to produce image attentions. And both of question and attended image features are used to predict the answers.

However, answering visual question requires not only the visual content information but also common knowledge about them, which can be too hard to directly learn from a limit number of images with only QA supervision. And comparatively few previous research worked on enriching the knowledge base when performing the VQA task. We are aware of three related papers. \cite{li2018vqa} firstly use annotated captions to build explanation dataset in an offline fashion then adopt a multi-task learning strategy which simultaneously learns to predict answers and generate explanation. Different with them, we use the captions as input which can provide richer feature during predicting. \cite{li2018tell} firstly generates captions and attributes with a fixed annotator and then use them to predict answers. Therefore, the captions they generated are not necessary related to the question and they also drop out the image features in the answer prediction process. \cite{rajani:naacl18} stacks auxiliary features to robustly predict answers. However, all the above do not utilize the complementarity between image captioning and the VQA task. Therefore, we propose to use captions to provide addition knowledge as well we use question to provide heuristics to the image captioner. \\

\noindent\textbf{Image Captioning}\\ 
Most of modern image captioning systems are attention based deep learning system  \cite{donahue2015long,karpathy2015deep,vinyals2015show}. With the help of large image description datasets \cite{chen2015microsoft}, those image captioning systems have shown remarkable results under automatic evaluation metrics. Most of them takes image embeddings from CNNs as inputs and build an attentional RNN ($i.e.$ GRU \cite{cho2014learning}, LSTM \cite{hochreiter1997long}) as language models to generate image captions without making prefixed decisions ($e.g.$ object categories). 

However, deep network systems still tend to generate similar captions with beam search strategy failing to diversely describe the images \cite{vijayakumar2016diverse}. In this work, rather than directly asking the system to mining diverse descriptions from images, we propose to providing more heuristics when generating the captions.

\section{Conclusion}
In this work, we explore the complementarity of the image captioning task and the VQA task. In particular, we present the joint system which generates question-related captions with question features as heuristics and uses the generated captions to provides additional common knowledge to help the VQA system. We produce more informative captions while outperform the current stat-of-the-art in terms of the VQA accuracy. Furthermore, we demonstrate the importance and necessity of including additional common knowledge besides the images in the VQA tasks and more heuristics in image captioning tasks. 
\bibliography{emnlp2016}
\bibliographystyle{acl_natbib_nourl}

\end{document}